\documentclass[letterpaper, 10pt, conference]{ieeeconf} 

\IEEEoverridecommandlockouts 

\usepackage{eso-pic} 
\usepackage[utf8]{inputenc} 
\usepackage[T1]{fontenc} 
\usepackage[colorlinks=true, allcolors=blue, bookmarks=true]{hyperref}
\usepackage{url} 
\usepackage{booktabs} 
\usepackage{amsfonts} 
\usepackage{amsmath}
\usepackage{nicefrac} 
\usepackage{microtype} 
\usepackage{sidecap} 
\usepackage{multirow}
\usepackage{graphicx}
\usepackage{bm}
\usepackage{amssymb,amstext}
\usepackage[nameinlink,capitalize]{cleveref}
\makeatletter
\let\NAT@parse\undefined
\makeatother
\usepackage[numbers]{natbib}
\usepackage{adjustbox}
\usepackage{lipsum}
\usepackage{multicol}
\usepackage{xcolor,colortbl}
\usepackage[font=small,labelfont=bf]{caption}
\usepackage{subcaption}
\usepackage{wrapfig}
\usepackage{footnote}
\usepackage{flushend} 
\let\labelindent\relax
\usepackage{enumitem}
\usepackage{lscape} 
\usepackage{pifont}
\usepackage{makecell}
\usepackage[flushleft]{threeparttable}

\makesavenoteenv{tabular}
\makesavenoteenv{table}

\definecolor{pltred}{rgb}{0.839, 0.153, 0.157}

\newcommand{\rom}[1]{\uppercase\expandafter{\romannumeral #1\relax}.}

\definecolor{Light}{RGB}{193, 237, 246}
\definecolor{Gray1}{gray}{0.0}
\definecolor{Gray2}{gray}{0.1}
\definecolor{Gray3}{gray}{0.25}
\definecolor{Gray4}{gray}{0.4}

\usepackage[ruled,vlined]{algorithm2e}
\renewcommand{\baselinestretch}{0.99}

\title{Non-Prehensile Throwing: A Reinforcement Learning Perspective}

\author{Abdullah Mustafa$^{1}$, Ryo Hanai$^{1}$, Ixchel G. Ramirez-Alpizar$^{1}$, Floris Erich$^{1}$, \\ Ryoichi Nakajo$^{1}$, Yukiyasu Domae$^{1}$ and Tetsuya Ogata$^{2}$
\thanks{$^{1}$A. Mustafa, R. Hanai, I. Ramirez-Alpizar, F. Erich, R. Nakajo, and Y. Domae are with the National Institute of Advanced Industrial Science and Technology (AIST), Japan
        {\tt\small \{am-mustafa, ryo.hanai, ixchel-ramirezalpizar, floris.erich, ryoichi-nakajo, domae.yukiyasu\}@aist.go.jp}}%
\thanks{$^{2}$T. Ogata is with the Graduate School of Fundamental Science and Engineering, Waseda University, Tokyo 169-8555, Japan and also with the National Institute of Advanced Industrial Science and Technology (AIST), Japan
        {\tt\small ogata@waseda.jp}}%
}

\usepackage{float}

\begin{document}

\makeatletter
\let\@oldmaketitle\@maketitle
\renewcommand{\@maketitle}{\@oldmaketitle
}
\makeatother

\maketitle
\IEEEpeerreviewmaketitle

\begin{abstract}
Robotic throwing enables fast object transport and extends a robot’s reachable workspace beyond traditional pick-and-place. While prehensile (grasp-based) throwing works well for graspable items, non-prehensile (grasp-free) throwing is better suited for large, heavy, and/or deformable objects. Existing approaches rely on model-based optimization with simplified contact models (e.g., dynamic grasping) and low-dimensional trajectory parameterizations, which limit solution quality and reachable workspace.
We propose a reinforcement learning approach that additionally leverages sliding and rolling contact modes and directly optimizes joint-space trajectories without analytical contact models or custom parameterizations. The Markov Decision Process (MDP) is formulated as a dynamical system that evolves the robot’s joint state conditioned on the throwing target, object model, and initial configuration. Joint-jerk trajectories are planned offline at a low control rate and upsampled into smooth, high-rate velocity commands for deployment. For sim-to-real transfer, we minimize the robot-dynamics gap through minimum-jerk system identification and train uncertainty-aware policies to mitigate object-modeling errors, particularly sensitivity to dynamic friction.
In simulation, the policy achieves 99\% success across thousands of configurations and generalizes to unseen objects. Sensitivity analysis shows robustness to mass uncertainty but high sensitivity to dynamic friction, consistent with the sliding-based release mechanism. Deployed zero-shot on a UR5e operating near its physical limits (5\,m/s end-effector velocity), our method throws diverse objects—including heavy (790\,g) and large ($20\times20\times28$\,cm) items—to targets up to 350\,cm distance or 180\,cm elevation, achieving a \textbf{97\%} real-world success rate.
\end{abstract}

\section{Introduction}

Throwing enables robots to transport objects rapidly and extend their effective workspace beyond quasi-static pick-and-place. As a result, robotic throwing has received increasing attention, with two main modes considered in the literature: prehensile (using a gripper) and non-prehensile (using a tray).

\textbf{Prehensile throwing} performs well for small objects~\citep{tossBot, tubeAcc}, where a trajectory is optimized to reach a desired end-effector pose and velocity before gripper release. Larger or heavier objects can be manipulated using oversized industrial grippers~\citep{3Dthrow} or bi-manual systems~\citep{bimanualThrow}, but these systems are difficult to coordinate at high velocities and less precise for smaller items. For deformable objects, prehensile throwing can induce grasp instability and uncertain release timing~\citep{tubeAcc}.

\begin{figure}[t]
\centering
\begin{adjustbox}{minipage=\linewidth-4pt,margin=0pt 0pt,bgcolor=gray,frame=1pt}
    \centering
\includegraphics[width=\linewidth, trim={0mm 0mm 0mm 0mm},]{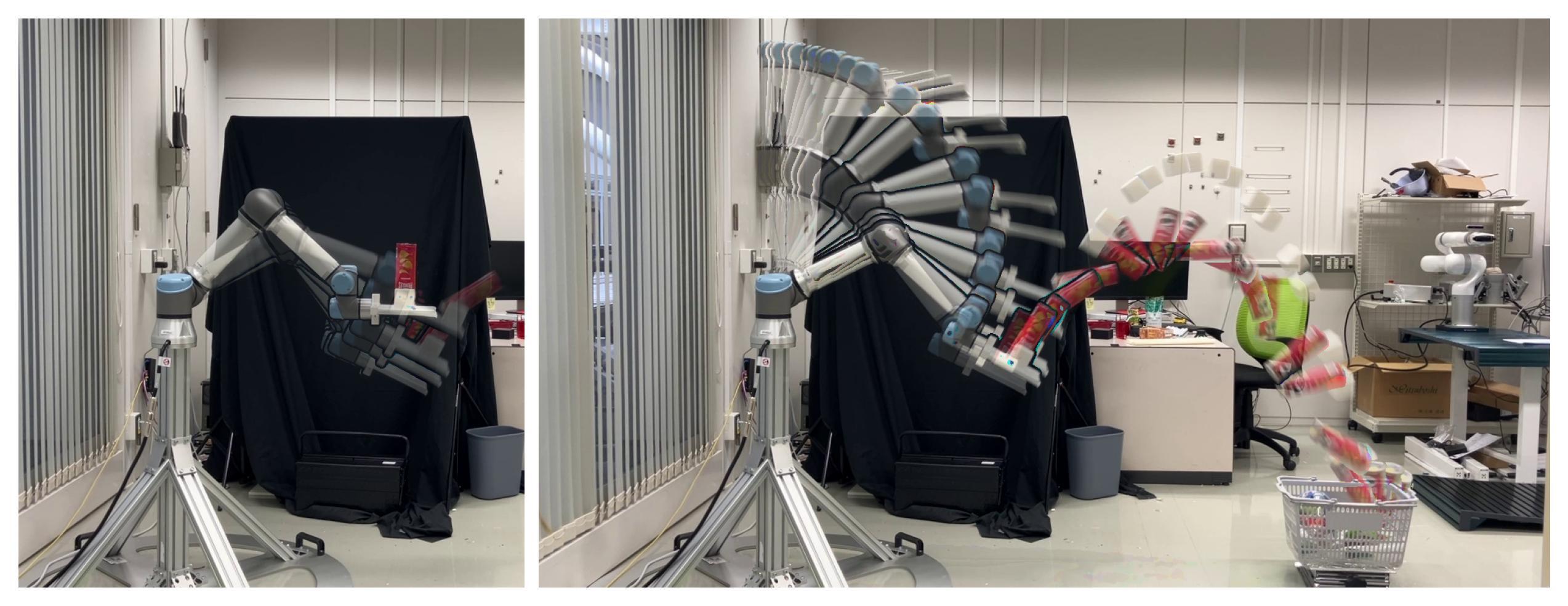}
\end{adjustbox}
\begin{adjustbox}{minipage=\linewidth-4pt,margin=0pt 0pt,bgcolor=gray,frame=1pt}
    \centering
\includegraphics[width=\linewidth, trim={0mm 0mm 0mm 0mm},]{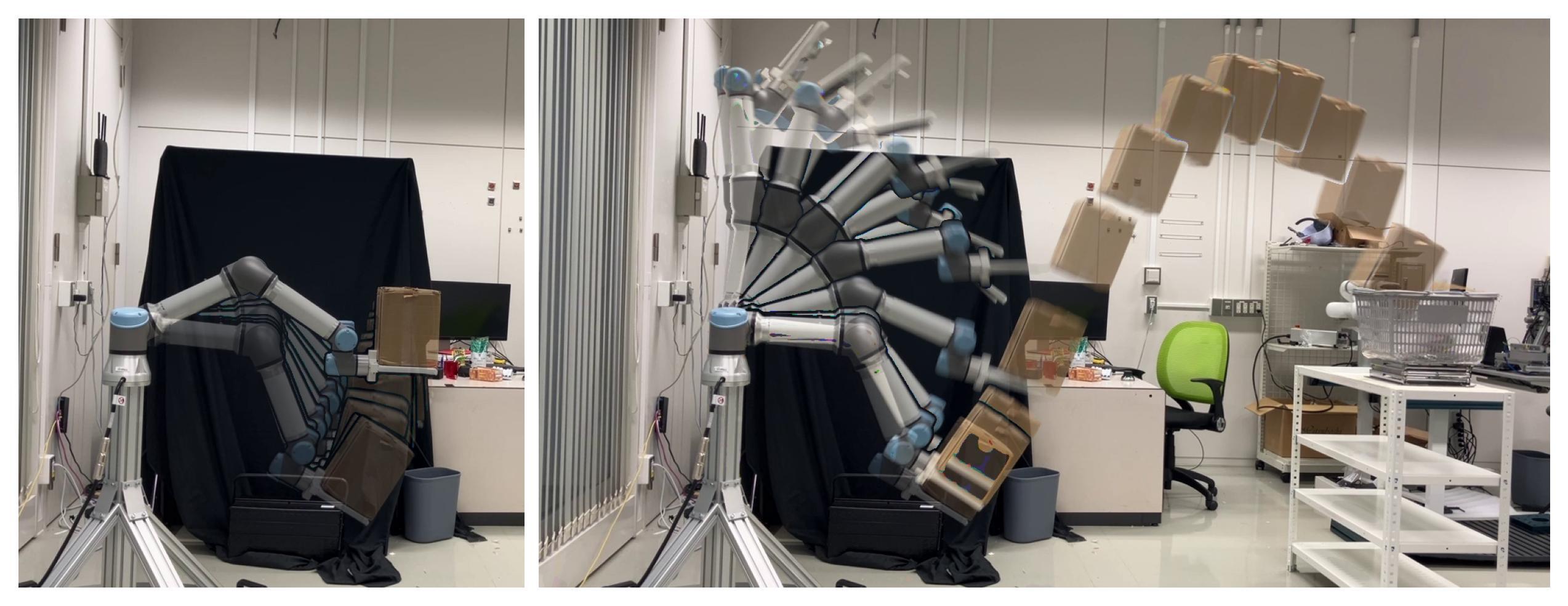}
\end{adjustbox}
\begin{adjustbox}{minipage=\linewidth-4pt,margin=0pt 0pt,bgcolor=gray,frame=1pt}
    \centering
\includegraphics[width=\linewidth, trim={0mm 0mm 0mm 0mm},]{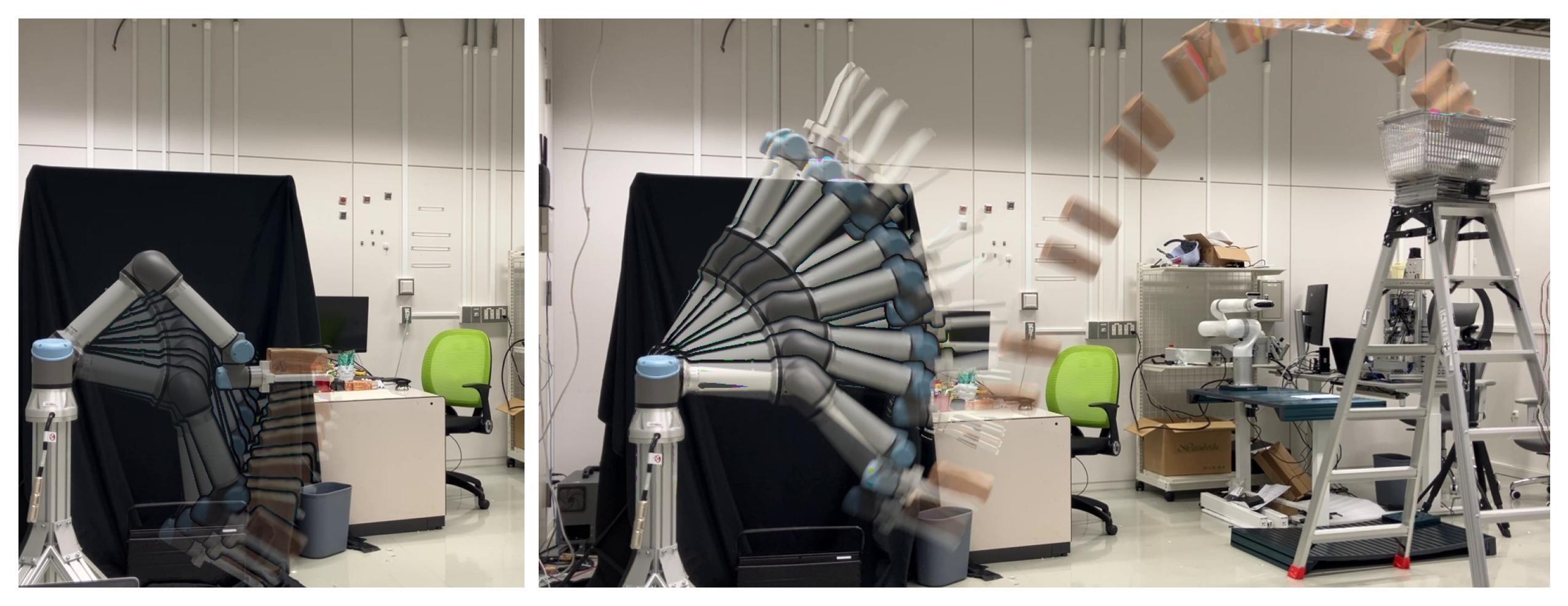}
\end{adjustbox}

\caption{\emph{NP-Throw} enables planar non-prehensile throwing of generic objects to distant and elevated targets. Its sliding-based release mechanism avoids the infeasible fast decelerations used by dynamic-grasping approaches. 
\textbf{Top} (200\,cm, 15\,cm): Multiple objects, including a tall chips can and a small cube. 
\textbf{Middle} (250\,cm, 105\,cm): Large box. 
\textbf{Bottom} (300\,cm, 185\,cm): Heavy wood block. 
Videos: \url{https://tinyurl.com/np-throwing}.}
\label{fig:teaser}
\end{figure}

\textbf{Non-prehensile throwing} offers an alternative that naturally accommodates large, heavy, or deformable objects. However, without grasp forces, the system must satisfy kino-dynamic constraints to avoid unintended sticking, sliding, or rolling. Prior work enforces these constraints through dynamic grasping~\citep{lynchDNP, NP-Throw-Lynch, NP-throw-planar, oneJointThrow}, slip-less rolling for spherical objects~\citep{ballThrowing}, or mechanically constrained setups~\citep{ballThrowingEdge}. While these approaches provide valuable modeling insight, their deployment relies on simplified contact models and low-dimensional trajectory parameterizations. Moreover, optimization is typically performed via sequential quadratic programming (SQP), which is sensitive to initialization and prone to local minima and solver failures—ultimately restricting performance and limiting applicability to a small set of carefully designed scenarios.

\textbf{RL perspective.} Reinforcement learning offers a complementary approach that does not require fixed contact-mode assumptions or hand-designed trajectory parameterizations. In a physics simulator, RL can discover additional hybrid sliding/rolling interactions and generate explicit joint-space trajectories by training over thousands of randomized environments—overcoming the expressiveness limits of analytical models and the brittleness of SQP solvers.

\textbf{This work.} We propose \emph{NP-Throw}, an RL-based non-prehensile throwing policy. The MDP is formulated as a dynamical system defined over implicit robot-state observations; the policy outputs joint-jerk commands at a low rate, which are upsampled into smooth velocity trajectories for deployment. To bridge the sim-to-real gap, we (i) calibrate robot dynamics via minimum-jerk system identification, and (ii) train uncertainty-aware policies to address object-modeling errors—especially sensitivity to dynamic friction. 

\textbf{Key results.} In simulation, the policy attains $\sim$99\% success across thousands of in-distribution configurations and generalizes effectively to unseen YCB objects~\citep{YCB}. Sensitivity analysis shows robustness to mass-estimation errors but high sensitivity to dynamic-friction mismatch, consistent with the policy's reliance on sliding-based release. When deployed on real hardware, \emph{NP-Throw} achieves an average success rate of \textbf{97\%} across five challenging objects and long-range targets (up to 350\,cm distance or 185\,cm elevation), operating near the robot’s physical limits (5\,m/s end-effector velocity). Sample planar throwing trajectories are shown in \cref{fig:teaser}

\textbf{Contributions.}
We present an RL formulation for planar non-prehensile throwing that learns hybrid contact behaviors without fixed contact-mode assumptions or trajectory parameterizations. We provide extensive simulation studies analyzing policy's performance, out-of-distribution generalization, and sensitivity to modeling errors. We introduce a practical sim-to-real pipeline—leveraging uncertainty-aware training and minimal dynamics tuning—that enables reliable zero-shot deployment on real hardware.

\textbf{Code:} \url{https://tinyurl.com/np-throwing}

\section{Related Works}
\textbf{\emph{Non-prehensile Throwing.}} 
To simplify trajectory optimization, prior work often adopts low-dimensional trajectory parameterizations—such as B-splines~\citep{lynchDNP, oneJointThrow}, radial basis functions~\citep{ballThrowing}, or polynomials~\citep{ballThrowingEdge}—which restrict the feasible trajectory space and can yield suboptimal solutions. To handle hybrid dynamics, most approaches assume a single contact mode (e.g., dynamic grasping~\citep{lynchDNP, NP-throw-planar, oneJointThrow}) or separately optimize motion primitives with predefined mode transitions~\citep{NP-Throw-Lynch, ballThrowing}, further limiting exploration. Experimental demonstrations are typically constrained to a single object type (e.g., cuboid or sphere) and specialized setups such as tilted air-hockey tables~\citep{NP-Throw-Lynch, NP-throw-planar, ballThrowing}.

\textbf{\emph{Prehensile Throwing.}}
To address the sensitivity of model-based optimization to initialization, several works use offline-generated datasets to improve and accelerate trajectory optimization. In~\citep{mobileThrow}, a dataset mapping release states to robot states is combined with a throwing manifold to obtain multiple \emph{good} initial guesses. Nearest-neighbor retrieval in landing or release space provides approximate trajectories, and failed executions are reused to reduce the sim-to-real gap~\citep{baxterThrow, throwflip}. Differentiable motion-manifold primitives are learned and  latent samplers are refined with real data to generate improved initial trajectories~\citep{MMPLEE, DA-MMP}. Extending these dataset-driven strategies to non-prehensile throwing inherits limitations of the underlying data-generation process—namely, low-dimensional parameterization and single-mode contact assumptions. The additional kino-dynamic constraints in non-prehensile settings also make trajectory generation for generic objects substantially more computionally expensive. Although random throwing trajectories bypass optimization and reduce dataset cost~\citep{baxterThrow, DTthrow}, they would fail to adequately cover the reachable workspace in non-prehensile scenarios.

\textbf{\emph{Learning-based Prehensile Throwing.}} 
End-to-end learning offers an alternative. Self-supervised residual policy~\citep{tossBot} corrects release pose and velocity directly on real hardware, while~\citep{dartBot} learns dart throwing via off-policy RL with constrained action paramterization. Simulation-trained throwing policies can transfer zero-shot to real systems~\citep{throwMoving}. However, reducing the action space to a release-state parameterization is insufficient for non-prehensile throwing, where the release state emerges from long-horizon contact-rich dynamics. Nonetheless, large-scale RL with explicit joint-space actions, as shown in dexterous manipulation~\citep{DexPBT}, has provided a viable framework for learning complex manipulation tasks.

\section{Problem Formulation}

\begin{figure}[t]
\centering
\includegraphics[width=0.8\linewidth]{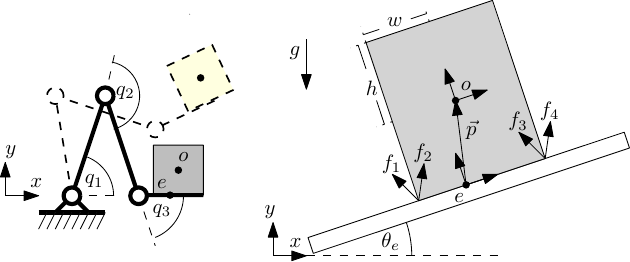}
\caption{Robot and object configurations for planar non-prehensile throwing.}
\label{fig:model}
\end{figure}

We address the problem of motion planning for non-prehensile throwing. Owing to equivariance in the throwing direction, we focus on planar throwing. Given the object model, the initial robot/object configurations, and a throwing target, a throwing trajectory is generated offline and executed without feedback on the object state. \cref{fig:model} illustrates the robot and object configurations. Without loss of generality, we consider a fixed-base robot with three coplanar revolute joints ($\bm{q}=[q_1,\,q_2,\,q_3]$). A uniform-density cuboid object is described by its mass $m$, height $2h$, width $2w$, and static/dynamic friction coefficients $\mu_s,\mu_k$ (collectively $\beta=[m,h,w,\mu_k,\mu_s]$). The planar target is specified by the horizontal distance $r_\mathrm{t}$ and elevation $z_\mathrm{t}$ in the world frame.

Next, we introduce a simplified dynamic-grasping model to (i) motivate the MDP design choices, (ii) interpret later key empirical findings, and (iii) highlight why analytical approaches struggle with hybrid contact dynamics.
\subsection{Dynamic-Grasping Throwing Model}

Following \citep{keepitupright}, we adopt the Newton--Euler balance ($\bm{w}_{\mathrm{GI}}+\bm{w}_{\mathrm{C}}=0$) to derive a simplified planar model, where $\bm{w}_{\mathrm{GI}}$ and $\bm{w}_{\mathrm{C}}$ denote the gravito-inertial and contact wrenches in the object frame. The object lies on a flat tray, with its pose relative to the end-effector given by $\bm{p}=[p_x,p_y]$. The end-effector state includes planar pose $\bm{x}_e=[x_e,y_e,\theta_e]$, velocities $\bm{v}_e$ and $\omega_e=\dot{\theta}_e$, and accelerations $\dot{\bm{v}}_e$ and $\dot{\omega}_e$.

Under sticking or sliding, the object’s rotational motion matches the end-effector ($\theta_o=\theta_e$, $\omega_o=\omega_e$, $\dot{\omega}_o=\dot{\omega}_e$). Its linear velocity and acceleration follow
\[
\bm{v}_o = \bm{v}_e + \omega_e \times \bm{p}, \qquad 
\dot{\bm{v}}_o=\dot{\bm{v}}_e + \dot{\omega}_e \times \bm{p} + \omega_e \times \dot{\bm{p}},
\]
yielding
\begin{equation}
\label{eq:gravitoInertial_wrench_planar}
\bm{w}_{\mathrm{GI}} =
\begin{bmatrix}
m(\ddot{x}_e + g\sin\theta_e - \ddot{\theta}_e p_y) \\
m(\ddot{y}_e - g\cos\theta_e + \ddot{\theta}_e p_x + \dot{\theta}_e\dot{p}_x) \\
I\ddot{\theta}_e
\end{bmatrix}.
\end{equation}

Contact force is approximated using two point contacts. Each force is expressed as a positive combination of friction-cone edge vectors: $\bm{f}_{i_l}=[-\mu,1]f_{i_l}$ and $\bm{f}_{i_r}=[\mu,1]f_{i_r}$ with $f_{i_l},f_{i_r}\ge 0$. The resulting wrench is
\begin{equation}
\label{eq:contact_wrench_planar}
\bm{w}_{\mathrm{C}} = [F_t,\,F_n,\,F_t h + \tau]^T,
\end{equation}
where normal contact force is $F_n = f_1+f_2+f_3+f_4,$ 
tangential contact force is $F_t = -\mu(f_1+f_3)+\mu(f_2+f_4),$
and contact moment is $\tau = (-f_1-f_2+f_3+f_4)w$.

Because $f_i \ge 0$, the following friction-cone constraints enforce anti-slip and anti-roll behavior:
\begin{equation}
\label{eq:friction_cone_stick}
\mu F_n \ge |F_t|,\qquad w F_n \ge |\tau|.
\end{equation}

\begin{figure}[t]
\centering
\includegraphics[width=0.8\linewidth]{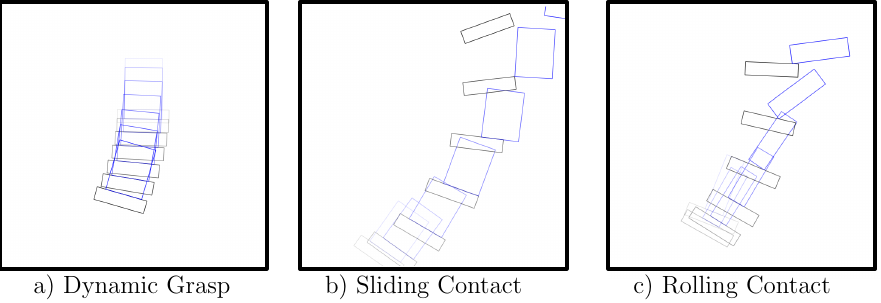}
\caption{Contact modes exhibited during \emph{NP-Throw}. The end-effector and object are shown in black and blue, respectively.}
\label{fig:contactModel}
\end{figure}

\textbf{\emph{Challenges of Analytical Models.}}
Throughout the trajectory (\cref{fig:contactModel}), the object can exhibit dynamic grasp ($f_{1,2,3,4}>0$), sliding ($f_{1,3}=0$ or $f_{2,4}=0$), rolling ($f_{1,2}=0$ or $f_{3,4}=0$), or mixed modes. The earlier simplified model assumes no rolling and fixed contact locations on the object—assumptions that break as soon as significant sliding occurs. Accurately modeling these hybrid transitions analytically and optimizing over them is therefore non-trivial. While trajectory-optimization methods~\citep{lynchDNP, NP-Throw-Lynch, NP-throw-planar, oneJointThrow} typically assume a single contact mode or prescribe a fixed sequence, our objective is to learn a policy that implicitly controls hybrid contact dynamics and the necessary mode transitions for optimal task performance.

\textbf{\emph{Insights for MDP Design.}}
This simplified formulation provides three key insights for policy design:
\begin{itemize}
\item The object state is determined by the end-effector trajectory and object model in absence of disturbances, 
\item A minimal object model consists of mass (inertia), geometry (width/height), and contact friction (static/dynamic).
\item Controlling the end-effector at the acceleration or jerk level is essential for smooth contact-mode transitions.
\end{itemize}

\section{Reinforcement Learning Perspective}\label{sec:method}
We first introduce the proposed Markov Decision Process (MDP) for offline motion planning, followed by the procedure for successful sim-to-real transfer.

\subsection{Markov Decision Process Formulation}
\label{sec:MDP}

The objective of reinforcement learning (RL) is to learn a policy $\pi^*$ that maps a state $s_t \in \mathcal{S}$ to an action $a_t = \pi^*(s_t) \in \mathcal{A}$, maximizing the expected discounted return over a finite horizon, $R_t = \mathbb{E}\!\left[\sum_{i=t}^{T}\gamma^{\,i-t} r_{i+1}\right],$
where $T$ is the horizon length and $\gamma \in [0,1)$ is the discount factor.
We consider a fully observed, deterministic MDP in which the initial state $s_0 \sim \mathcal{S}_0$ evolves according to a desired dynamical system:
\[
\bm{s}_{t+1}=f(\bm{s}_t,\,a_t)=f(\bm{s}_t,\,\pi^*(s_t)).
\]

\begin{figure}[t]
\centering
\includegraphics[width=\linewidth]{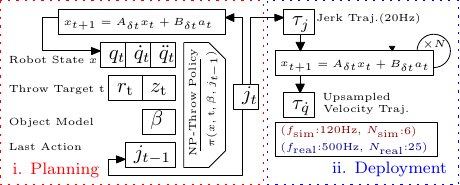}
\caption{Throwing trajectory planning using the \emph{NP-Throw} policy. Given an object model $\beta$, throwing target $\mathrm{t}=(r_\mathrm{t},z_\mathrm{t})$, and initial joint configuration $x$, the jerk trajectory is generated offline by recursively evaluating the policy. The resulting trajectory is upsampled into high-rate velocity commands for execution in simulation and on real hardware.}
\label{fig:arch}
\end{figure}

The transition function $f$ is defined solely over the robot state. Although the object state is not explicitly included, in the absence of disturbances it is implicitly determined by the robot’s trajectory and the object model. The overall architecture is illustrated in \cref{fig:arch}.

\subsubsection{\textbf{Initial State Distribution} $\mathcal{S}_0$}
The initial state includes the robot state $x$, object model parameters $\beta$, and the target $(r_\mathrm{t},z_\mathrm{t})$. Table~\ref{tab:ranges} summarizes the sampling distributions used for throwing-configuration randomization. These ranges were selected empirically based on our robot, tray design, and object set, and should be adapted appropriately for other systems.
The object is modeled as a cuboid, and we randomize its mass, width, height, and static/dynamic friction coefficients. The initial robot configuration is randomized such that the tray remains horizontal and points along the positive $x$-direction (\cref{fig:model}). All joint velocities, accelerations, and jerks are initialized to zero. The throwing target is sampled to span the reachable distance and elevation enabled by hardware limits (5\,m/s maximum end-effector velocity for UR5e robot).
To ensure full observability, the object is always initialized at the center of the tray, and its center of mass (COM) is assumed to coincide with the geometric center. Additional randomization may be incorporated towards further robustness.

\begin{table}[t]
\caption{Sampling ranges for randomizing the initial state $s_0$.}
\centering
\begin{threeparttable}
{\begin{tabular}{l|c|c}
\toprule
\rowcolor{gray!30}
\textbf{Property} & \textbf{Unit} & \textbf{Distribution} \\
\midrule
\rowcolor{gray!15}
\multicolumn{3}{c}{\textbf{Object Model} $\beta$} \\
Width & cm & $\mathcal U(5,\,20)$ \\
Height & cm & $\mathcal U(5,\,25)$ \\
Mass & kg & $\mathcal U(0.1,\,1.0)$ \\
Static/Dynamic Friction & -- & $\mathcal U(0.2,\,1.0)$ \\
\rowcolor{gray!15}
\multicolumn{3}{c}{\textbf{Throwing Target}} \\
Distance & cm & $\mathcal U(150,\,350)$ \\
Elevation & cm & $\mathcal U(0,\,200)$ \\
\rowcolor{gray!15}
\multicolumn{3}{c}{\textbf{Joint Configuration} $q$} \\
Joint 1 ($q_1$) & rad & $-\pi/4 \pm \mathcal U(-\pi/4,\,\pi/4)$ \\
Joint 3 ($q_3$) & rad & $-\pi/4 \pm \mathcal U(-\pi/4,\,\pi/4)$ \\
Joint 2 ($q_2$)\tnote{a} & rad & $q_2 = -(q_1 + q_3)$ \\
\bottomrule
\end{tabular}}
\begin{tablenotes}
\small
\item \textsuperscript{a} Ensures a horizontal tray at initialization for UR5e robot.
\end{tablenotes}
\end{threeparttable}
\label{tab:ranges}
\end{table}

\subsubsection{{Observation Space}}
The observation space comprises the robot state (position, velocity, acceleration) $\bm{x}_t=[\bm{q}_t,\, \dot{\bm{q}}_t,\, \ddot{\bm{q}}_t]$, the object model $\beta=\{m,\, w,\, h,\, \mu_s,\, \mu_k\}$, the throwing target $(r_\mathrm{t},z_\mathrm{t})$, and the last action $a_{t-1}$. We also ablated the use of extended histories $\bm{x}_{t-k:t}$ and $\bm{a}_{t-k-1:t-1}$, which did not improve performance. The robot state $\bm{x}_t$ is updated recursively given the jerk action $\dddot{\bm{q}}_t$, without accessing ground-truth simulated or real state.
In absence of disturbances, the object state can be implicitly inferred by the policy knowing the end-effector trajectory and object model. The object model $\beta$ follows the planar mechanics derived in the previous section. To improve learning stability and performance, we adopt an asymmetric actor–critic architecture: the critic receives privileged information, including end-effector and object states (pose and velocity), as well as a binary indicator of whether the object has been released (i.e., zero contact force with the tray). This indicator, $i_{\mathrm{thrown},t}$, is used in the reward and is important for accurate value estimation.

\subsubsection{{Action Space}}
While a high control frequency can yield smooth trajectories and improved tracking, it also increases trajectory length and slows simulation and learning. We therefore design the action space to ensure smoothness with minimal horizon length for task success. Specifically, we use jerk-based actions ($a_t=\dddot{\bm{q}}$) together with up-sampling for robust trajectory tracking. Given the joint jerk, the robot joint state $\bm{x}_t=[\bm{q}_t,\, \dot{\bm{q}}_t,\, \ddot{\bm{q}}_t]$ is updated as
\begin{equation}
\bm{x}_{t+1} = {\bm{A}}\bm{x}_t + {\bm{B}}\bm{a}_t, 
\label{eq:action}
\end{equation}
where
\begin{align*}
{\bm{A}} &= \begin{bmatrix}
\bm{I}_3 & \delta t\bm{I}_3 & (1/2)\delta t^2\bm{I}_3 \\
\bm{0}_3 & \bm{I}_3 & \delta t\bm{I}_3 \\
\bm{0}_3 & \bm{0}_3 & \bm{I}_3
\end{bmatrix}, &
{\bm{B}} &= \begin{bmatrix}
(1/6)\delta t^3\bm{I}_3 \\ (1/2)\delta t^2\bm{I}_3 \\ \delta t\bm{I}_3
\end{bmatrix}.
\end{align*}

Here, $\bm{I}_3$ is the 3$\times$3 identity matrix, and $\delta t$ is the integration step (with control frequency $f = 1/\delta t$). The policy operates at a lower frequency $f_{\pi}$, while simulation and hardware run at higher frequencies $f_{\mathrm{sim}} = N_{\mathrm{sim}} f_{\pi}$ and $f_{\mathrm{real}} = N_{\mathrm{real}} f_{\pi}$. Holding the jerk action constant over \[\delta t_{\pi} \;=\; N_{\mathrm{sim}}\,\delta t_{\mathrm{sim}} \;=\; N_{\mathrm{real}}\,\delta t_{\mathrm{real}}, \]
we generate an up-sampled joint-velocity trajectory with different discretizations for simulation and real deployment (see \cref{fig:arch}). The joint velocity and acceleration are clipped to remain within their prescribed limits,
$\dot{\bm{q}}_{\min} \le \dot{\bm{q}} \le \dot{\bm{q}}_{\max}$ and 
$\ddot{\bm{q}}_{\min} \le \ddot{\bm{q}} \le \ddot{\bm{q}}_{\max}$.

\subsubsection{{Reward Formulation}}
The reward function encourages fast and accurate throwing and comprises live-time, dense shaping, sparse success, and regularization terms.

\begin{itemize}[noitemsep, wide, labelwidth=!, labelindent=0pt] 
\item  \emph{Live–Thrown} ($r_{L}=i_{\mathrm{live},t}\, i_{\mathrm{thrown},t}$): During early training, these two binary terms dominates, encouraging longer episodes ($i_{\mathrm{live},t}$) and early object release ($i_{\mathrm{thrown},t}$). Using only $i_{\mathrm{live},t}$ led to unnecessary object holding.
 
\item \emph{Position Error} ($r_{p}=\lVert \bm{p}_{O,t}-\bm{p}_{\mathrm{target}}\rVert_2$): This dense term is crucial for guiding initial exploration, which stalls when only sparse rewards are used. The error is the $\ell_2$ distance between the object position $\bm{p}_{O,t}$ and the target $\bm{p}_{\mathrm{target}}$.

\item \emph{Velocity to Target} 
($r_{v_{\mathrm{goal}}}=\Big\langle \tfrac{\bm{v}_{O,t}}{\lVert \bm{v}_{O,t}\rVert_2}, \tfrac{\bm{p}_{\mathrm{target}}-\bm{p}_{O,t}}{\lVert \bm{p}_{\mathrm{target}}-\bm{p}_{O,t}\rVert_2}\Big\rangle$): This term encourages movement toward the target via the dot product between the normalized object velocity and the normalized direction-to-target vector.

\item \emph{Success} ($r_{\mathrm{success}}=i_{\mathrm{success},t}\,\mathcal{K}(r_p; a)$): This is the primary objective for the bin-throwing task. The binary success indicator $i_{\mathrm{success},t}$ is $1$ when the object is within a $0.25\,\mathrm{m}$ threshold (half the real basket width) of the target. Accuracy is further encouraged by a smooth kernel $\mathcal{K}(r_p; a)=\frac{1}{e^{a r_p}+e^{-a r_p}}$, 
which increases exponentially as $r_p$ decreases.

\item \emph{Regularization Penalties.}
To obtain smooth and efficient trajectories, we penalize joint acceleration $r_{\ddot{q}}=\lVert \ddot{\bm{q}}_{t}\rVert_2$ and jerk $r_{\dddot{q}}=\lVert \dddot{\bm{q}}_{t}\rVert_2$. To damp motion post-release, we heavily penalize joint velocity after the throw via $r_{\dot{q}}=\lVert \dot{\bm{q}}_{t}\rVert_2 \cdot i_{\mathrm{thrown},t}$. These penalties are scaled relative to the mean of $r_{\mathrm{success}}$ avoiding excessive regularization early in training phases.

\end{itemize}
The total reward is a weighted sum of the above terms, with empirically-tuned weights ($
w_L=10,\quad w_p=-1, w_{v_{\mathrm{goal}}}=10, w_{\mathrm{success}}=100,
w_{\dot{q}}=-100, w_{\ddot{q}}=-0.1, w_{\dddot{q}}=-0.01$).

\subsubsection{{Episode Termination}}
During training, episodes are terminated early upon failure to prevent low-quality data from entering the training buffer. Failure includes the object contacting the ground ($p^{z}_{O,t}<0$). A time-limit termination is also applied at the fixed horizon.

\subsubsection{{Implementation Details}}
We use the NVIDIA IsaacLab framework~\citep{mittal2023orbit} for efficient parallel data collection in NVIDIA Isaac Sim and train the policy with RL-Games~\citep{rl-games} using Proximal Policy Optimization (PPO)~\citep{schulman2017proximal}. 
Our physical platform is a Universal Robots UR5e (maximum reach $\!\!<\!1$\,m), operated up to an end-effector velocity of $5$\,m/s (with safety limits disabled). Control is issued via \texttt{ur-rtde}~\citep{rtde}, a Real-Time Data Exchange wrapper for UR robots.

The policy is a feed-forward MLP with five hidden layers $[1024,\,512,\,256,\,128,\,64]$ and GeLU activations. The trajectory horizon is $3.2$\,s, corresponding to $64$ steps at a $20$\,Hz policy frequency. The up-sampled velocity trajectory is tracked at $120$\,Hz in simulation and $500$\,Hz on hardware (UR5e maximum). Additional architectural details are provided in our code release. Training across $4096$ environments for $1000$ epochs (about $2.62\times10^{8}$ steps) took a little over one hour on a single NVIDIA RTX~4090 GPU notebook.

\subsection{Sim-to-Real Transfer}
The simulation-trained policy can be reliably transferred to real hardware by minimizing the sim-to-real gap in both robot and object dynamics.

\subsubsection{\textbf{Robot Dynamics Gap}}
Prior to policy training, we ensure that commanded joint-velocity trajectories produce consistent behavior across simulation and hardware.  
To minimize this robot dynamics gap, we execute a set of minimum-jerk cyclic trajectories with varying amplitudes and frequencies on the physical system and record the resulting joint positions and velocities.  
The same trajectories are then executed in simulation across $N$ parallel environments (4096), each with different actuator stiffness and damping parameters.  
The optimal parameters are selected by minimizing a weighted sum of joint-position and joint-velocity errors between the simulated and recorded real trajectories.  
A second refinement stage samples additional parameter sets in a neighborhood around the optimized values to further reduce the error.

\subsubsection{\textbf{Object Dynamics Gap}}
Unlike the robot dynamics gap—which is minimized once—the object dynamics must be adjusted for each new object to maximize success. Although parameters such as size, mass, and static friction can often be estimated with reasonable accuracy, these estimates may still lead to failures due to the sim-to-real gap. To minimize the required tuning effort, we follow the steps below:

\begin{enumerate}
    \item We conduct a sensitivity analysis (\cref{fig:robust}) to identify object properties most sensitive to modeling errors. This analysis reveals strong robustness to mass uncertainty but high sensitivity to dynamic-friction errors.
    \item We train uncertainty-aware policies that explicitly account for friction-model estimation errors  (\cref{fig:random}).
    \item For each object, we minimally tune only the dynamic-friction coefficient to maximize success at a chosen throwing target, and then reuse that value across all remaining targets.
\end{enumerate}

Thanks to the robustness of the uncertainty-aware policy, we empirically find that a single object model can apply to multiple objects while maintaining high success rates.

\section{Evaluation}\label{sec:experiments}
We first conducted extensive simulation experiments to evaluate the policy’s in-distribution \emph{performance} (\cref{fig:perform}) and its out-of-distribution \emph{generalization} (Table~\ref{tab:generalization}). We performed \emph{ablation} studies (\cref{fig:ablations}) to assess the impact of key design choices. Prior to real-world deployment, we carried out a \emph{sensitivity} analysis (\cref{fig:robust}) to evaluate robustness to modeling errors and then trained an \emph{uncertainty-aware} policy (\cref{fig:random}) to mitigate sensitivity to dynamic-friction mismatch. Finally, real-world evaluation across five objects and twelve throwing targets achieved an average success rate of 97\% (Table~\ref{tab:real}).

\textbf{\emph{Note regarding baselines}}:
Due to the absence of open-source implementations for model-based non-prehensile throwing, we could not perform fair comparisons against such methods. Using the ``Upright'' implementation~\citep{keepitupright}—originally developed for non-prehensile transportation—we attempted dynamic-grasping-based throwing. While partial success was achievable, solver failures were frequent and required careful tuning of the initial configuration, release pose/velocity, and solver hyperparameters (e.g., horizon length, cost weights, slack variables). Combining the strengths of RL and MPC represents an interesting direction for future work.

\subsection{Simulation Results}

\subsubsection{Performance Analysis}\label{res:perform}
\begin{figure*}[t]
\centering
\includegraphics[width=1.0\linewidth]{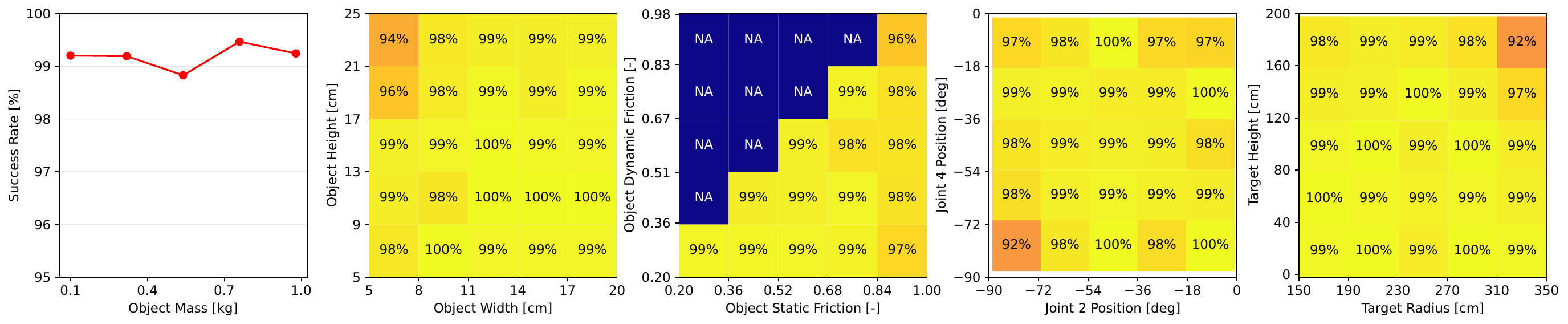}
\caption{Performance analysis of \emph{NP-Throw} policy for in-distribution configurations. The success rate is shown as a line plot for object mass and as 2D histograms for pairs of configuration variables. Entries with dynamic friction higher than static friction are not applicable (N/A). The policy achieves consistently high success rates except under extreme combinations of object size, joint state, and throwing target.}
\label{fig:perform}
\end{figure*}

We evaluated policy performance using 12{,}288 configurations (4096 per seed across three random seeds), sampled according to Table~\ref{tab:ranges}. The average success rate was approximately 99\%. \cref{fig:perform} shows how performance varies with different configuration parameters.

Performance degradation was observed primarily for extreme configurations that rendered the throw physically infeasible. For instance, when the robot starts with the first two links nearly vertical, the end-effector cannot generate sufficient acceleration to reach distant and elevated targets—particularly for objects with high height-to-width ratios. While object mass had limited influence on performance, higher static friction slightly reduced success rate. Empirically, we observed that sliding tends to dominate the motion, and larger friction coefficients increase the required end-effector accelerations to overcome frictional resistance. These findings suggest that surfaces with lower static and dynamic friction are advantageous for real-world deployment of our policy.

\subsubsection{Policy Generalization} \label{res:general}
Beyond the training distribution, we evaluated the policy’s generalization to unseen objects and throwing configurations, as summarized in Table~\ref{tab:generalization}. When evaluating cylindrical objects, we observed only a minor drop in performance, suggesting that both cuboids and cylinders exhibit similar contact behavior under our planar throwing policy. Our selection of YCB objects~\citep{YCB} included cuboid (“wood block”), cylindrical (“chips”), and generic shapes (“power drill”), all achieving high success rates. Performance degradation primarily occurred for configurations that are physically infeasible for the robot. Objects with large height-to-width ratios tend to roll or violate anti-roll constraints, particularly for distant or elevated targets. For unseen object models, the policy adapts reasonably well to lower masses and heights. Very low friction causes rapid sliding and insufficient object acceleration, hindering long-distance throws. Conversely, high mass or high friction induces larger friction forces, delaying sliding and limiting the achievable gravito-inertial acceleration. For tall objects, satisfying anti-rolling constraints while producing high acceleration may not be feasible. Finally, limits on maximum joint velocity, acceleration, and jerk impose corresponding limits on the set of reachable target positions for a given object model.

\begin{table}[t]
\caption{Evaluation of policy generalization to unseen objects and throwing configurations. The policy generalizes well to cylindrical objects and YCB items~\citep{YCB}, while performance degrades for infeasible throwing configurations. Success rates below 75\% are highlighted in red. The ``\textcolor{red}{-Short}'' label denotes aligning the object's short edge along the $x$-axis, increasing its height-to-width ratio.}
\centering
\begin{tabular}{l|c||l|c}
\toprule
\rowcolor{gray!30}
\textbf{Object} & \textbf{S.R. [\%]} & \textbf{Object} & \textbf{S.R. [\%]} \\
\midrule
Cuboids [Training] & 98.97 & Cylinders & 98.54 \\
\midrule
\rowcolor{gray!15}
\multicolumn{4}{c}{\textbf{YCB Objects}} \\
WoodBlock & 99.83 & Chips & 97.68 \\
Spam & 99.88 & Pitcher & 99.51 \\
\midrule
CrackerBox & 99.90 & CrackerBox-\textcolor{red}{Short} & 96.44 \\
MustardBottle & 99.85 & MustardBottle-\textcolor{red}{Short} & 87.04 \\
BleachCleanser & 99.44 & BleachCleanser-\textcolor{red}{Short} & {79.49} \\
PowerDrill & 95.65 & PowerDrill-\textcolor{red}{Short} & 89.28 \\
\midrule
\rowcolor{gray!15}
\multicolumn{4}{c}{\textbf{Object Model}} \\
Mass [0.01--0.1] kg & 98.97 & Mass [1--3] kg & \cellcolor{red!20}{71.92} \\
Friction [0.01--0.2] & 87.06 & Friction [1--3] & \cellcolor{red!20}{53.93} \\
Height [1--5] cm & 96.44 & Height [25--40] cm & 79.22 \\
\midrule
\rowcolor{gray!15}
\multicolumn{4}{c}{\textbf{Target Position}} \\
Elevation [2--2.5] m & 81.54 & Elevation [2.5--3] m & \cellcolor{red!20}{30.83} \\
Distance [3.5--4] m & \cellcolor{red!20}{71.63} & Distance [4--4.5] m & \cellcolor{red!20}{13.13} \\
\bottomrule
\end{tabular}
\label{tab:generalization}
\end{table}

\subsubsection{Ablation Studies} \label{res:ablation}
\begin{figure*}[t]
\centering
\includegraphics[width=1.0\linewidth]{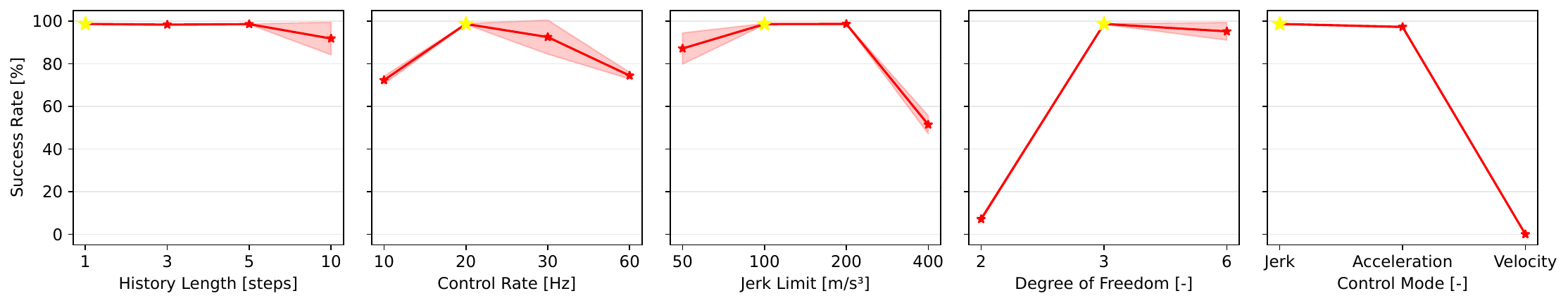}
\caption{Ablation study across different design axes. Success rates (mean and standard deviation) are computed over 4096 configurations and 3 random seeds. Our proposed policy configuration achieves the highest success rate, highlighted by a yellow star.}
\label{fig:ablations}
\end{figure*}

For each ablated design choice, we trained a policy for 500 epochs using three random seeds and evaluated it over 4096 in-distribution configurations. \cref{fig:ablations} compares success rates across the different design axes. In general, policy optimization becomes more challenging as the observation space size (history length), episode length (control rate), or action-space dimensionality (degrees of freedom) increases. Larger input spaces slow convergence and reduce success rates under a fixed compute budget (500 epochs). Using only the most recent joint state $\bm{x}_t$ provided sufficient observability while keeping the policy input compact. While higher control rates increase temporal resolution, they also lengthen the episode horizon, making credit assignment more difficult. Conversely, too low a control rate (e.g., 10~Hz) restricts the feasible trajectory space. Similarly, a low jerk limit cannot generate sufficient acceleration for throwing, whereas a large jerk limit requires more exploration to discover optimal actions. Lower jerk limits are also preferable in real-world deployment due to safer motions. Using all 6~DOF of the robot showed no benefit in our planar throwing task and introduced additional action dimensions that slightly reduced performance. The unexpectedly poor performance of a 2-DOF system is an interesting finding and warrants further investigation. Finally, both jerk-based and acceleration-based actions can induce the necessary contact-mode transitions, whereas velocity-based actions lead to more aggressive switching. While learning at high control rates with velocity actions may produce smoother trajectories, it incurs significantly higher computational cost.

\subsubsection{Policy Sensitivity to Modeling Uncertainty} \label{res:sens}
\begin{figure*}[t]
\centering
\includegraphics[width=1.0\linewidth]{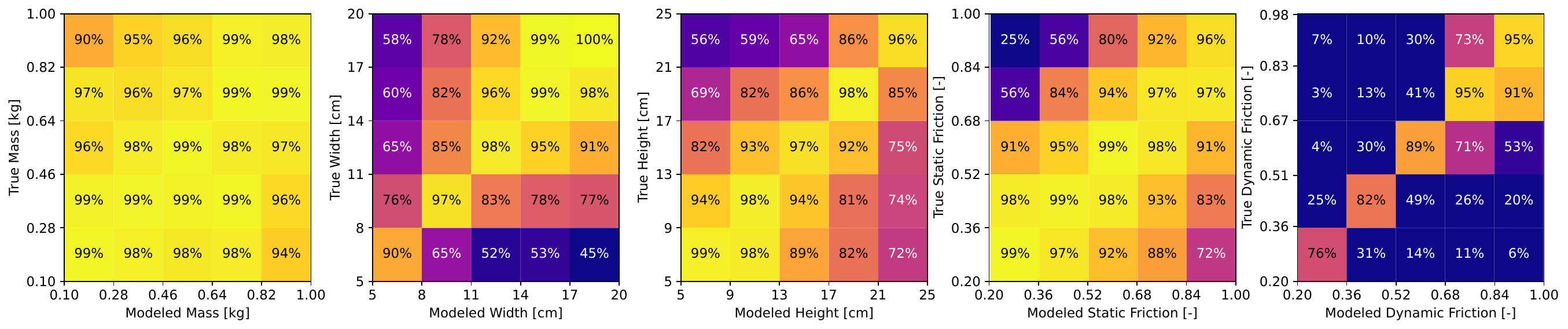}
\caption{Evaluation of policy robustness to modeling uncertainties, measured by success rate. As the estimated object model (x-axis) deviates from the true model (y-axis), performance is minimally affected by mass uncertainty but is most sensitive to dynamic-friction errors.}
\label{fig:robust}
\end{figure*}

Estimating the true object model is nontrivial for real-world deployment due to the sim-to-real dynamics gap. To minimize the required tuning effort, we evaluate the policy’s sensitivity to modeling errors. For each component of $\beta$ (e.g., mass), we fix the remaining parameters to their true values and randomly sample an estimate of that component from the distributions in Table~\ref{tab:ranges}. Aggregated results over 4096 configurations and three random seeds are shown in \cref{fig:robust}.

When perturbing the modeled mass relative to the true mass, the policy maintains near-perfect success rates, indicating insensitivity to mass uncertainty. From Eqs.~\eqref{eq:gravitoInertial_wrench_planar}, \eqref{eq:contact_wrench_planar}, and \eqref{eq:friction_cone_stick}, the anti-slip and anti-roll conditions are
${\mu} F_n \ge |F_t|$ and $w F_n \ge |\tau|$. Substituting the contact-wrench terms $F_n$, $F_t$, and $\tau$ with their gravito-inertial equivalents shows that the mass $m$ cancels in both inequalities. Accordingly, provided the end-effector tracks the commanded jerk/velocity trajectory accurately, the object experiences the same contact modes and release state. Nevertheless, the required joint torques do scale with mass and are bounded by hardware limits.

Object-size accuracy is more critical for thin (small true width) and tall (large true height) objects, for which anti-rolling constraints are harder to satisfy under uncertainty. The policy is moderately robust to errors in static friction but highly sensitive to dynamic friction: except for extremely high true static friction, underestimating friction tends to produce trajectories that overcome sticking; however, the release state under sliding depends strongly on the dynamic friction coefficient. Unlike prior work~\citep{lynchDNP, NP-throw-planar, oneJointThrow} that emphasizes dynamic grasping (more sensitive to static friction), our policy predominantly leverages sliding contact, making dynamic friction the key factor for accurate release.

\begin{figure}[t]
\centering
\includegraphics[width=0.9\linewidth]{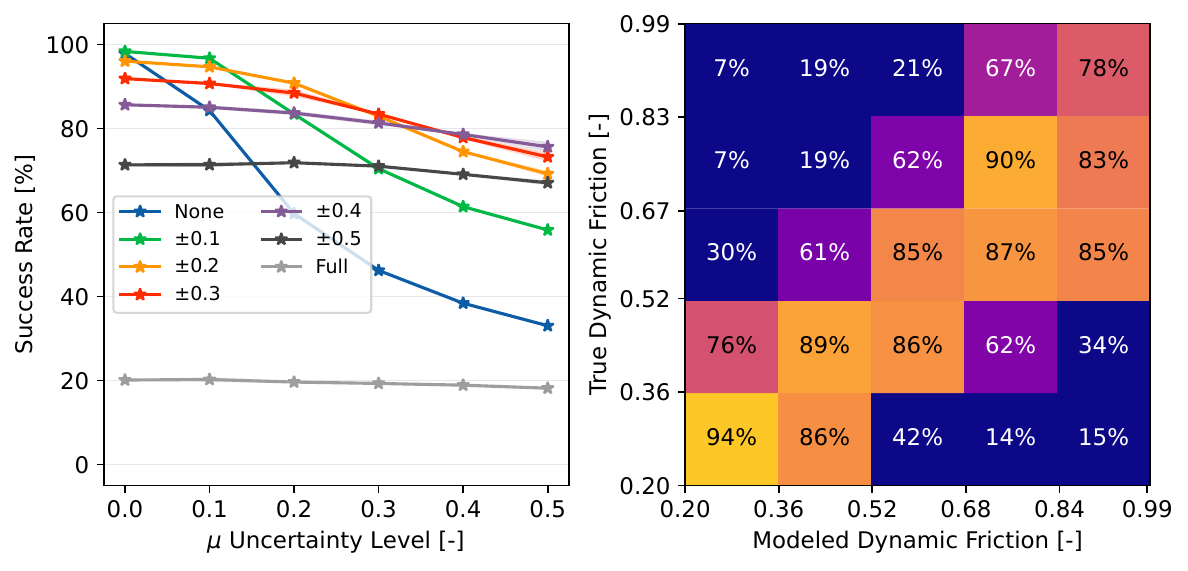}
\caption{\textbf{Left}: Performance of uncertainty-aware policies starting from zero uncertainty training (``None'') to fully random (``Full'') for different levels of friction uncertainty. \textbf{Right}: Sensitivity to dynamic friction for the policy trained with $\pm 0.3$ friction noise.}
\label{fig:random}
\end{figure}

Given this sensitivity, we train uncertainty-aware policies before real deployment. For each episode, we add a uniform noise ($\pm \nu$) to the true static and dynamic friction values. \cref{fig:random} compares success rates for policies trained under increasing friction uncertainty. The policy trained with perfect models (``None'') exhibits a sharper performance drop as uncertainty grows, whereas uncertainty-aware policies degrade more gracefully. Conversely, training with overly large noise (``Full'') prevents learning effective throws except for nearby targets. The friction-sensitivity plots further show improved robustness for uncertainty-aware training, which is preferable in practice.

\subsection{Real-World Evaluation}
\label{res:real}

\begin{figure}[t]
\centering
\includegraphics[width=0.25\linewidth, angle=270]{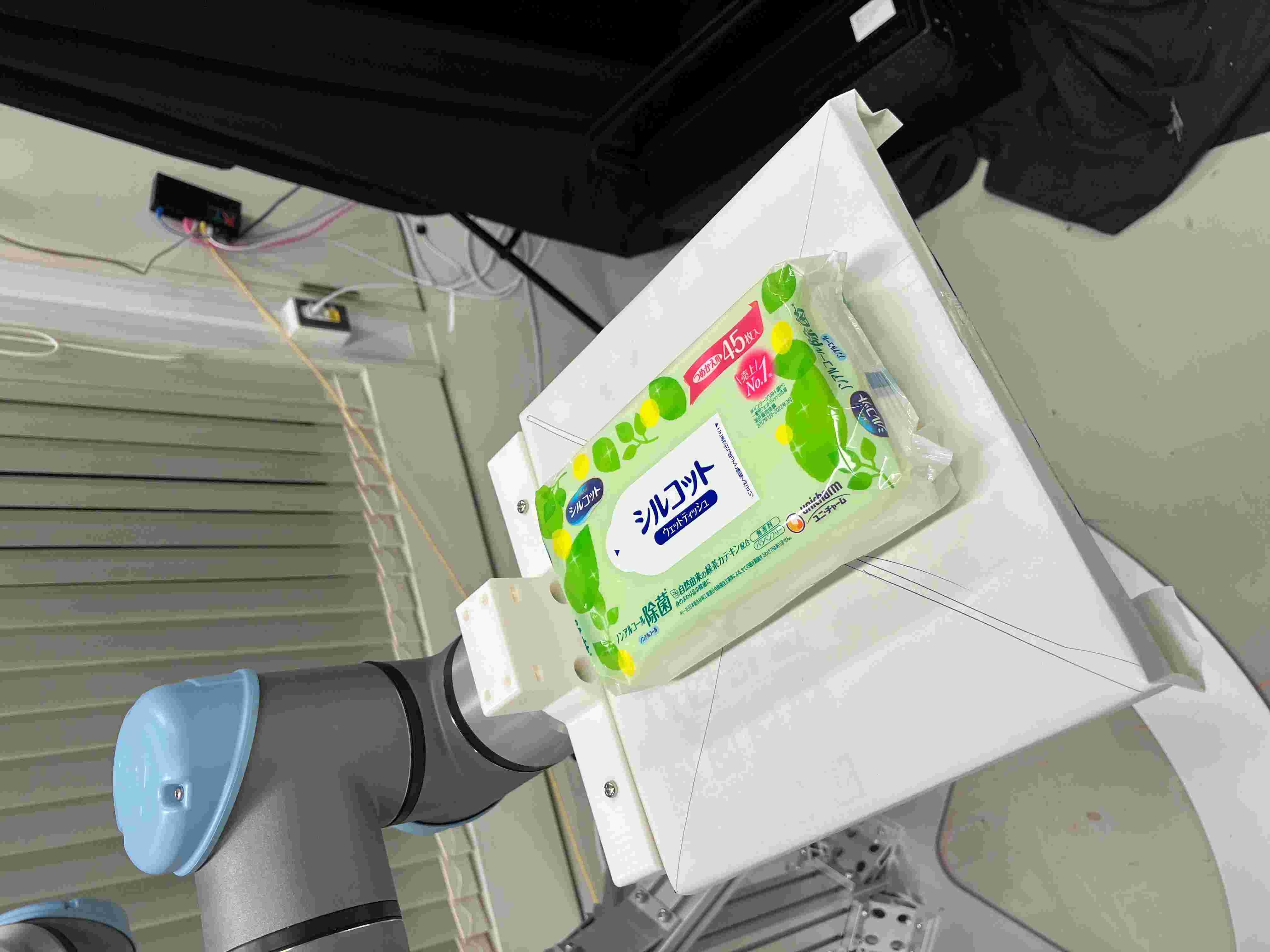}\hfill
\includegraphics[width=0.25\linewidth, angle=270]{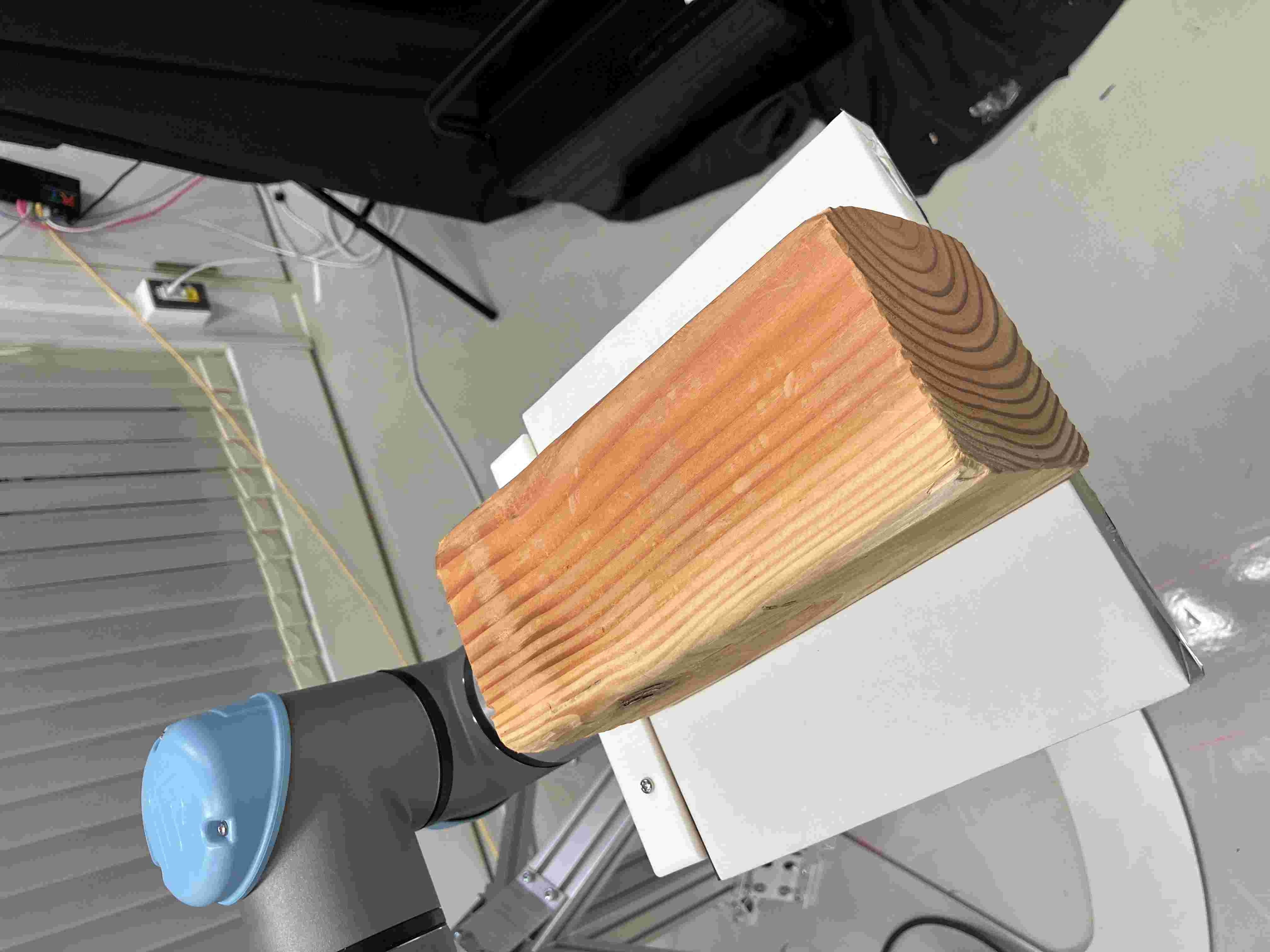}\hfill
\includegraphics[width=0.25\linewidth, angle=270]{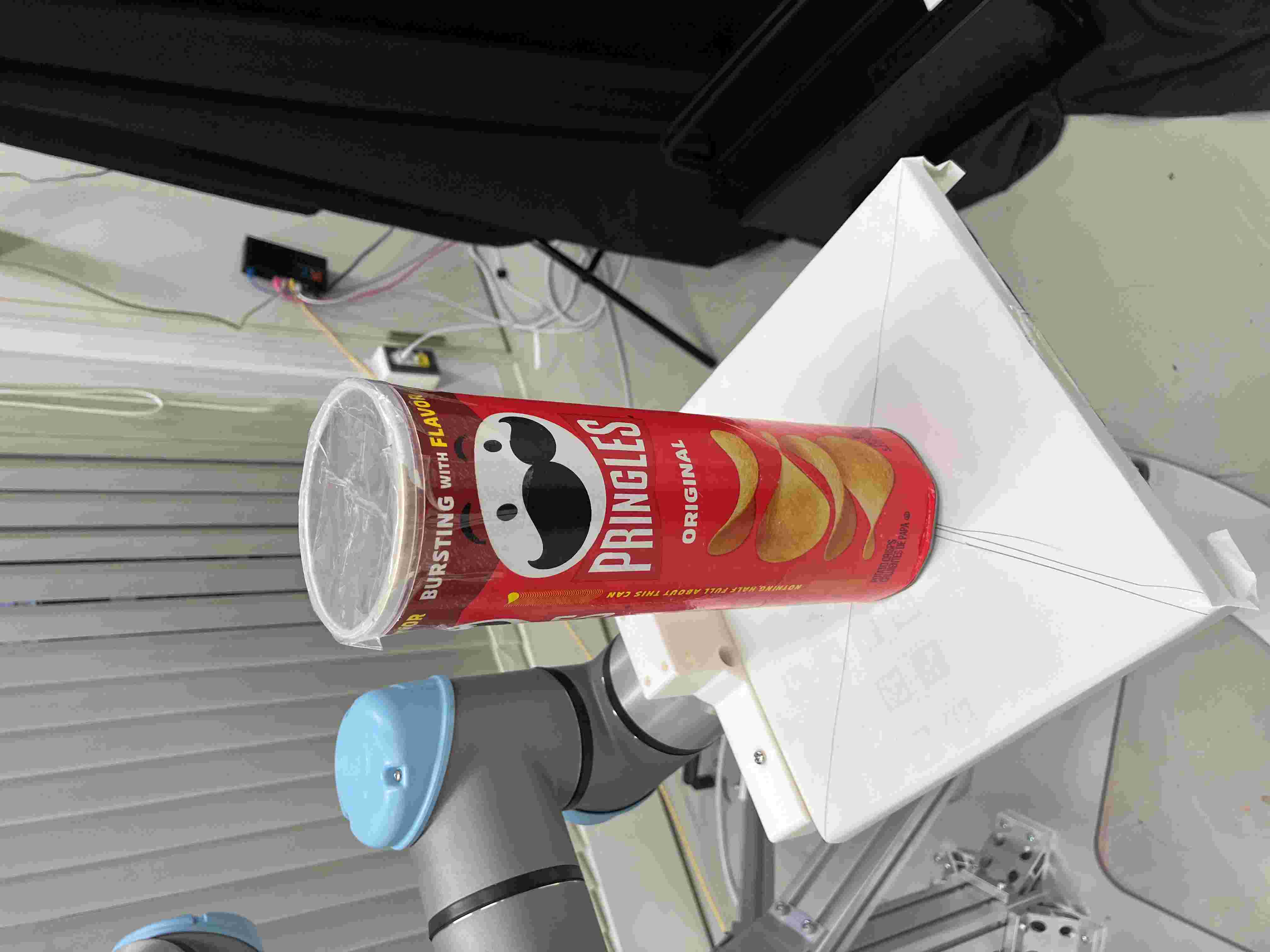}\hfill
\includegraphics[width=0.25\linewidth, angle=270]{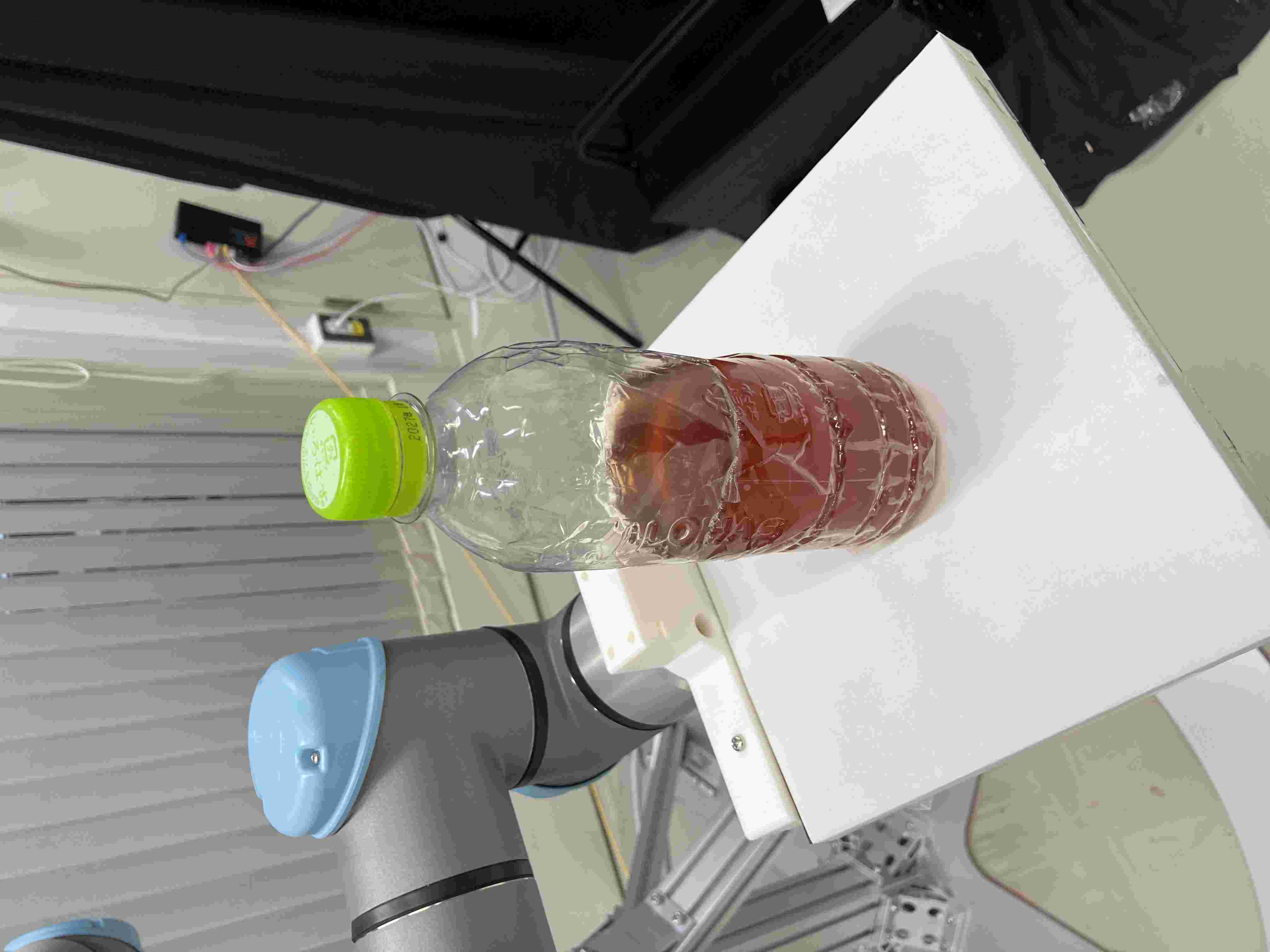}\hfill
\includegraphics[width=0.25\linewidth, angle=270]{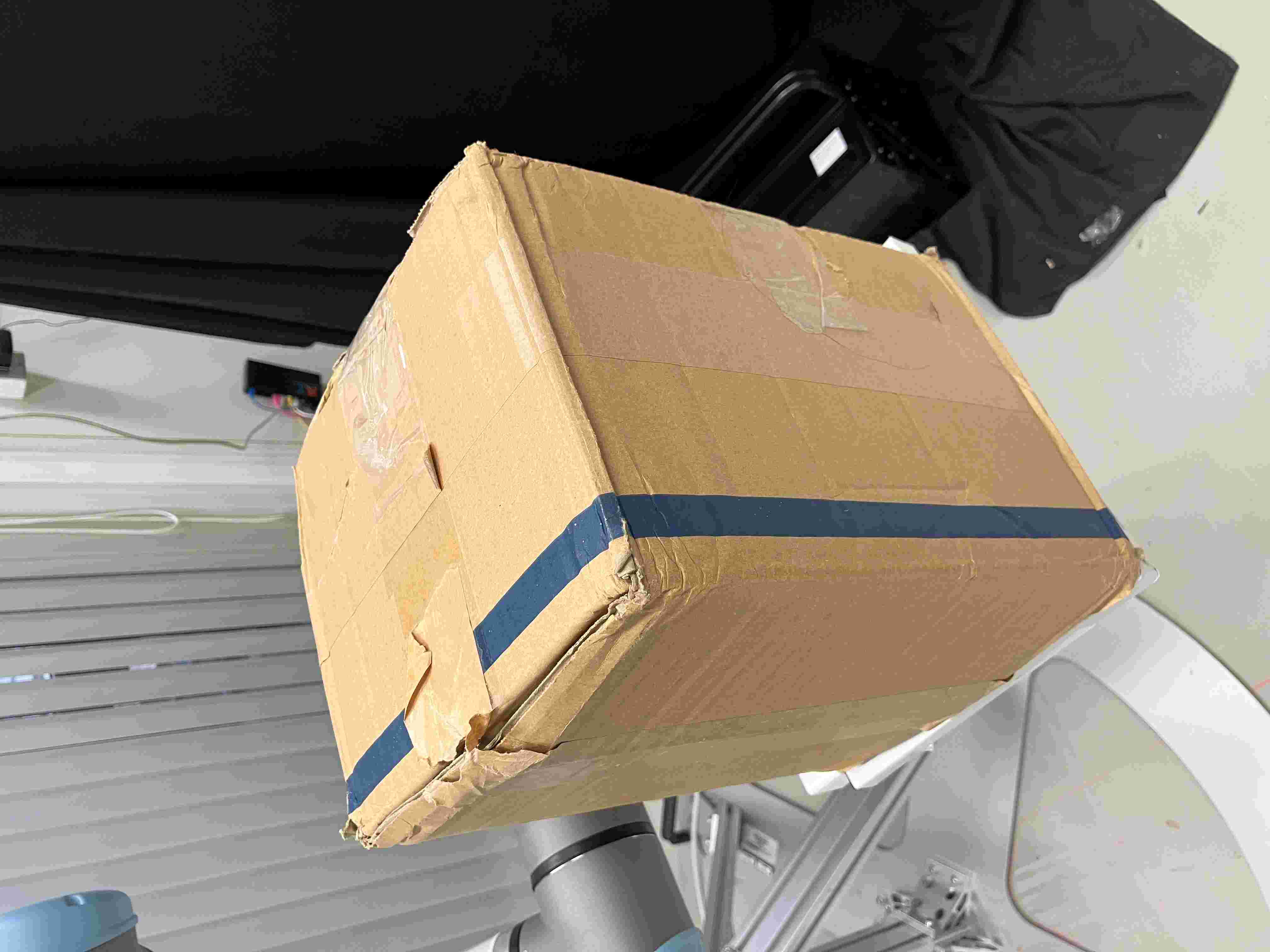}
\caption{Real-world evaluation objects: \emph{deformable} tissue box (\#1), \emph{heavy} $790$\,g wood block (\#2), \emph{tall cylindrical} chips (\#3), \emph{partially filled} bottle (\#4), and \emph{large} $20\times20\times28$\,cm box (\#5).}
\label{fig:objects}
\end{figure}

\emph{\textbf{Evaluation Setup.}}
We selected five objects with distinct physical properties as illustrated in \cref{fig:objects}. Throwing targets consisted of four distances (200, 250, 300, 350\,cm) and three elevations (15, 105, 185\,cm), resulting in 60 combinations ($5\times4\times3$). For each configuration, the policy was first evaluated in simulation across 64 randomized initial joint configurations; the first 10 successful trajectories were then executed on hardware. Following simulation findings (\cref{fig:perform}), we placed a sheet of paper on the tray to reduce the static friction introduced by the original rubber surface. The static and dynamic friction coefficients were tuned empirically using the wood block at the $(250\,\mathrm{cm},\,105\,\mathrm{cm})$ target, resulting in $\mu_s = 0.5$ and $\mu_k = 0.2$, which were applied uniformly across all objects and targets. Leveraging the policy’s robustness to mass-model errors (\cref{fig:robust}), we approximated all object masses as 0.5\,kg. Object size was estimated using an axis-aligned bounding box. For robust deployment, we used the policy trained with $\pm0.3$ friction uncertainty (\cref{fig:random}).

\emph{\textbf{Results.}}
Table~\ref{tab:real} reports an average success rate of 97\% across the 60 evaluated combinations (excluding the invalid configuration). At the $(350\,\mathrm{cm},\,185\,\mathrm{cm})$ target, all objects required release velocities that caused the policy-generated end-effector trajectory to dip too low—an artifact of the robot stand not being modeled during training—leading to collisions. In addition, the maximum end-effector velocity (M.EE.V.) exceeded the hardware limit (5\,m/s), making real deployment unsafe. For other extreme targets (e.g., $(300\,\mathrm{cm},\,185\,\mathrm{cm})$), feasible trajectories could not be generated for high height-to-width ratio objects (e.g., the chips can).  Across the remaining targets, failures were primarily due to small placement inaccuracies (object not perfectly centered on the tray) or modeling errors that affected release timing—effects that are amplified at long distances. The average throwing duration was approximately $1$\,s, and when combined with efficient picking (e.g., conveyor-based feeding), the system demonstrates high throughput potential.

\textbf{\emph{Note on policy robustness.}}
The simulation robustness trends in \cref{fig:robust} were validated on hardware: trajectories optimized for a single object could be reused across a wide range of other objects while maintaining high success rates. As illustrated in \cref{fig:teaser}, two distinct objects could even be thrown simultaneously using the same planned trajectory.

\begin{table}[t]
\caption{Real-world evaluation for the objects in \cref{fig:objects}. The policy achieves an average \textbf{97\%} success rate across all objects and targets. Extreme targets (e.g., 350\,cm distance at 185\,cm elevation) failed (marked \textcolor{red}{F}) due to infeasible end-effector velocities and collisions with the robot base. We also report aggregated statistics ($\mu\!\pm\!\sigma$) for the maximum end-effector velocity (M.EE.V.) and throwing time (Throw T.) for each target.}
\centering
\begin{threeparttable}
\begin{tabular}{c|c|c c c c c|c|c|}
\toprule
$r_\mathrm{t}$ & $z_\mathrm{t}$ & \multicolumn{5}{c}{\textbf{Real Success Rate [/10]}} & \textbf{M.EE.V.} & \textbf{Throw T.}\\
\cmidrule{3-7}
[cm] & [cm] & \multicolumn{1}{c}{\textbf{\#1}} & \multicolumn{1}{c}{\textbf{\#2}} & \multicolumn{1}{c}{\textbf{\#3}} & \multicolumn{1}{c}{\textbf{\#4}} & \multicolumn{1}{c}{\textbf{\#5}} & [m/s] & [ms] \\
\midrule
\multirow{3}{*}{200} & 15  & 10 & 10 &  10 & 10 & 10 & 3.45$\pm$0.02 &  614$\pm$39 \\
                     & 105  &  10 &  10 & 10 & 10 & 10 & 3.87$\pm$0.07 &  701$\pm$49 \\
                     & 185 & 10 & 10 &  10 &  10 & 10 & 4.53$\pm$0.05 &  1141$\pm$50 \\
\midrule
\multirow{3}{*}{250} & 15  &  10 &  9 &  9 & 10 &  10 & 4.05$\pm$0.09 &  745$\pm$68 \\
                     & 105  &  10 &  10 &  10 & 10 &  10 & 4.07$\pm$0.06 &  1005$\pm$57 \\
                     & 185 &  10 &  10 &  10 & 9  &  10 & 4.70$\pm$0.04 & 1263$\pm$63 \\
\midrule
\multirow{3}{*}{300} & 15  &  10 &  8 &  9 & 10 &  10 & 4.10$\pm$0.07 &  1112$\pm$64 \\
                     & 105  &  10 &  10 &  10 & 9 &  8 & 4.39$\pm$0.02 &  1178$\pm$44 \\
                     & 185 &  10 &  10 &  \textcolor{red}{F} & \textcolor{red}{F}  &  \textcolor{red}{F} & 4.97$\pm$0.04 & 1390$\pm$59 \\
\midrule
\multirow{3}{*}{350} & 15  &  9 & 10 &  10 &  9 & 10 & 4.42$\pm$0.04 & 1265$\pm$51 \\
                     & 105  & 10 & 8 & \textcolor{red}{F} &  \textcolor{red}{F} &  \textcolor{red}{F} & 4.62$\pm$0.02 & 1226$\pm$48 \\
                     & 185 & \textcolor{red}{F} & \textcolor{red}{F} & \textcolor{red}{F} & \textcolor{red}{F} & \textcolor{red}{F} & 5.18$\pm$0.04 & 1455$\pm$50 \\
\bottomrule
\end{tabular}
\end{threeparttable}
\label{tab:real}
\end{table}

\section{Conclusion}
We presented a reinforcement-learning perspective on non-prehensile throwing that addresses key limitations of model-based approaches. In simulation, we extensively evaluated performance, generalization, and sensitivity, identifying dynamic friction as the pivotal parameter for our predominantly sliding-based release. With uncertainty-aware training, the simulation-trained policy transferred to hardware, achieving an average success rate of 97\%.

\textbf{Limitations \& Future Work.}
Beyond planar throwing, extending to full 3D throwing~\citep{3Dthrow} and increasing control DoF~\citep{tubeAcc} could yield more flexible strategies, particularly in cluttered environments. Rather than relying on manual and imprecise object modeling—especially for friction—future work will investigate autonomous dynamics identification with an emphasis on minimizing the sim-to-real gap. Another two complementary directions are (i) improving robustness to generic objects with irregular shape, nonuniform density, or non-planar contact, and (ii) increasing throwing precision for known object models. Finally, because heavy, long-distance throws can induce high-impact collisions, impact-aware throwing and complementary catching policies~\citep{throwCatch} are important for safer deployment.

{\small
\bibliographystyle{IEEEtranN}
\bibliography{BR}
} 

\end{document}